%% file: main.tex
\documentclass[10pt,twocolumn,letterpaper]{article}

\usepackage{cvpr}              
\PassOptionsToPackage{shortlabels,inline}{enumitem}
\usepackage{float}
\usepackage{algorithmic}
\usepackage{graphicx}
\usepackage{float}
\usepackage{textcomp}
\usepackage{xcolor}
\usepackage[table]{xcolor}
\usepackage{enumitem}
\usepackage{makecell}

\usepackage{booktabs}
\usepackage{siunitx}
\usepackage{caption}
\input{preamble}
\definecolor{cvprblue}{rgb}{0.21,0.49,0.74}
\definecolor{lightash}{RGB}{230, 230, 230}
\usepackage[pagebackref,breaklinks,colorlinks,allcolors=cvprblue]{hyperref}

\def\paperID{3} 
\def\confName{CVPR}
\def\confYear{2026}

\title{Firebolt-VL: Efficient Vision-Language Understanding \\ with Cross-Modality Modulation}

\author{Quoc-Huy Trinh$^{1, *}$, Mustapha Abdullahi$^{1, *}$, Bo Zhao$^{1,\dagger}$, Debesh Jha$^{2,\dagger}$\\
$^{1}$Aalto University \quad $^{2}$University of South Dakota\\
$^{*}$ Equal Contribution
$^{\dagger}$ Corresponding Author
}

\begin{document}
\maketitle
\input{sec/0_abstract}    
\input{sec/1_intro}
\input{sec/2_related}

\input{sec/3_method}

\input{sec/4_experiments}
\input{sec/5_conclusion}

{
    \small
    \bibliographystyle{ieeenat_fullname}
    \bibliography{main}
}


\end{document}

%% file: sec/0_abstract.tex
\begin{abstract}
Recent advances in multimodal large language models (MLLMs) have enabled impressive progress in vision-language understanding, yet their high computational cost limits deployment in resource-constrained scenarios such as personal assistants, document understanding, and smart cameras. Most existing methods rely on Transformer-based cross-attention, whose quadratic complexity hinders efficiency. Moreover, small vision-language models often struggle to precisely capture fine-grained, task-relevant visual regions, leading to degraded performance on fine-grained reasoning tasks that limit their effectiveness in the real world. To address these issues, we introduce Firebolt-VL, an efficient vision-language model that replaces the Transformer-based decoder with a Liquid Foundation Model (LFM) decoder. To further enhance visual grounding, we propose a Token-Grid Correlation Module, which computes lightweight correlations between text tokens and image patches and modulates via the state-space model with FiLM conditioning. This enables the model to selectively emphasize visual regions relevant to the textual prompt while maintaining linear-time inference. Experimental results across multiple benchmarks demonstrate that Firebolt-VL achieves accurate, fine-grained understanding with significantly improved efficiency. Our model and code are available at: \url{https://fireboltvl.github.io}

\end{abstract}

%% file: sec/1_intro.tex
\section{Introduction}

Multimodal Large Language Models (MLLMs) have rapidly advanced vision-language capabilities, achieving strong performance on tasks such as image captioning~\cite{li2023blip, liu_visual_2023}, visual question answering (VQA)~\cite{antol_vqa_2015, liu_visual_2023, zhu2023minigpt}, and optical character recognition (OCR)~\cite{wei_deepseek-ocr_2025,park_hierarchical_2024}. Recent state-of-the-art models such as LLaVA~\cite{llava,llava15,li2024llava}, IDEFICS~\cite{laurenccon2023obelics}, OpenFlamingo v2~\cite{openflamingo}, MiniGPT-4~\cite{zhu2023minigpt, chen2023minigpt}, Chameleon~\cite{team2024chameleon}, InternVL~\cite{internvl}, Qwen-VL~\cite{bai2023qwen}, and FastV~\cite{chen2024image} underscore the growing importance of MLLMs in real-world applications. Nevertheless, deploying MLLMs remains challenging in resource-constrained settings due to their high computational and memory demands. Improving efficiency is therefore critical for enabling low-latency and on-device multimodal interaction, achieving seamless integration of linguistic and visual reasoning in practical technologies.

Most recent MLLMs are built upon Transformer-based Large Language Models (LLMs), which exhibit quadratic computational complexity with respect to the input sequence length~\cite{vaswani2017attention, gu2024mamba}. As a result, inference is often inefficient on resource-constrained devices, and latency increases sharply for long-context multimodal inputs. Therefore, improving LLM efficiency is crucial to enable faster inference and facilitate the deployment of MLLMs in low-resource environments.

\begin{figure}[t]
    \centering
    \setlength{\belowcaptionskip}{-5pt}
    \begin{subfigure}[t]{0.58\columnwidth}
        \centering
        \includegraphics[width=\linewidth]{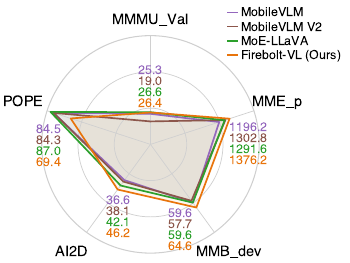}
        \caption{Task performance.}
        \label{fig:radar_mllm}
    \end{subfigure}\hspace{-0.06\columnwidth}
    \begin{subfigure}[t]{0.44\columnwidth}
        \centering
        \includegraphics[width=\linewidth]{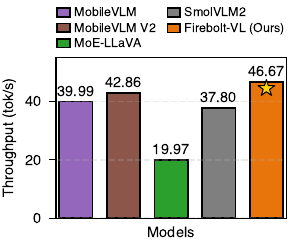}
        \caption{Throughput efficiency.}
        \label{fig:eff_throughput}
    \end{subfigure}
    \caption{Performance and latency efficiency of Firebolt-VL on multiple tasks compared with state-of-the-art baselines. Firebolt-VL is both competitive and efficient, demonstrating strong generalization across diverse tasks.}
    \label{fig:bench_eff_onecol}
\end{figure}

To alleviate the compute and memory overhead of MLLMs, several efficient architectures have been proposed, including Kosmos-2~\cite{peng2023kosmos}, MobileVLM~\cite{mobilevlm}, MobileVLM V2~\cite{mobilevlmv2}, MoE-LLaVA~\cite{llava-moe}, LLaVA-Phi~\cite{llavaphi}, and SmolVLM2~\cite{smolvlm}. These models either leverage lightweight language backbones or incorporate Mixture-of-Experts (MoE) mechanisms~\cite{moe} to reduce the number of active parameters and computational cost. Although such approaches have shown promising results on relatively simple benchmarks, such as image captioning and VQA, they still face two major challenges. First, small Transformer-based architectures exhibit quadratic computational complexity and limited capacity to model long-range dependencies. Second, there is a lack of precision when attending to task-relevant visual regions, which often leads to failures in handling fine-grained or detail-oriented questions that require rich visual representations.

In this work, we address these limitations by introducing a token-grid correlation and modulation mechanism, implemented as the Cross-Modal Modulator (CMM). This module fuses visual grid representations with instruction text tokens to emphasize task-relevant visual cues. By doing so, the CMM enables the model to attend more effectively to fine-grained and detail-oriented information, enhancing its ability to reason over complex visual inputs. Furthermore, we explore the integration of the Liquid Foundation Model (LFM)~\cite{hasani2023liquid, liquid_LFM} with our proposed CMM for an efficient MLLM. From all of our proposed approaches, we design \textbf{Firebolt-VL}, an efficient MLLM for broad vision-language understanding evaluated on tasks including image captioning, VQA, chart understanding, and fine-grained visual reasoning, as shown in Figure~\ref{fig:bench_eff_onecol}.

\noindent In summary, our main contributions are fourfold:

\begin{enumerate}[label=(\arabic*), itemsep=1ex, leftmargin=*, labelsep=1em, align=left, listparindent=1.5em, wide=0pt]
    \item We introduce \textbf{Firebolt-VL}, a novel MLLM that integrates the Liquid Foundation Model (LFM)~\cite{hasani2023liquid, liquid_LFM} for efficient sequence modeling, significantly reducing computational cost while maintaining strong multimodal reasoning performance.
    \item We propose a \textbf{Cross-Modal Modulator} mechanism that fuses visual grid representations with text tokens. This enables more precise attention to task-relevant regions and improves the model's capacity for fine-grained visual understanding.
    \item We conduct extensive experiments on multiple benchmarks, including image captioning, VQA, and OCR. Results demonstrate that Firebolt-VL achieves competitive or superior performance compared to existing efficient MLLMs, while substantially improving inference efficiency and scalability.
    \item We release the source code and pretrained model to promote transparency and encourage further research in the development of efficient MLLMs.
\end{enumerate}


%% file: sec/2_related.tex
\section{Related Work}

\subsection{Multimodal Large Language Model}

Multimodal large language models (MLLMs) have become a central research direction due to their wide applicability in document understanding, smart cameras, and virtual assistants. Recent advancements in state-of-the-art architectures~\cite{llava,laurenccon2023obelics,openflamingo,zhu2023minigpt,chen2023minigpt,llava15,li2024llava,team2024chameleon,internvl,bai2023qwen,chen2024image} have achieved remarkable progress in visual understanding and text generation, bringing MLLMs closer to practical, real-world deployment. 

Despite these advancements, the computational demands of modern Vision–Language Models (VLMs) remain a major barrier. To deal with this challenge, early efforts such as MobileVLM~\cite{mobilevlm} and MobileVLM V2~\cite{mobilevlmv2} reduce the computational burden by employing lightweight Mobile-LLaMA backbones for text generation. Subsequent approaches, including MoE-LLaVA~\cite{llava-moe} and LLaVA-Phi~\cite{llavaphi}, adopt Mixture-of-Experts (MoE)~\cite{moe} to activate only a fraction of parameters during inference, thereby eliminating redundant computation. More recently, SmolVLM2~\cite{smolvlm} introduced a language backbone combined with pixel-shuffle and inner-patching strategies to reduce the number of visual tokens to improve efficiency.

While these models show promising performance and increasing adoption, they still rely heavily on Transformer-based architectures whose attention mechanism incurs quadratic time and memory complexity. This fundamental limitation restricts their ability to scale to long-context inputs and prevents truly lightweight, real-time deployment. To address this limitation, the design of Firebolt-VL integrates a Liquid-based~\cite{hasani2023liquid} language model, which is leveraged by the Liquid Foundation Model (LFM) decoder~\cite{liquid_LFM}. This enables forward passes in linear-time complexity and significantly improves the overall efficiency of vision-language modeling.

\subsection{Cross-modal Integration}
In recent works, most VLMs introduce cross-modal alignment through a simple linear projection layer, which maps visual features into a joint embedding space shared with the language encoder. While effective for large-scale models with strong visual backbones, this strategy becomes problematic for compact VLMs, whose vision encoders possess limited representational capacity, often resulting in weak or unstable alignment.

To improve alignment quality in smaller models, several enhanced strategies have been proposed. Dense Connector~\cite{yao2024dense} enriches the visual representation by aggregating multi-level features from earlier layers. Align-KD~\cite{alignkd} leverages knowledge distillation to transfer cross-modal alignment cues from larger teacher models, thereby strengthening the alignment of compact VLMs. Building on this direction, Align-GPT~\cite{aligngpt} introduces a hierarchical alignment scheme that learns multiple alignment levels during pre-training and adaptively fuses them during instruction tuning to support diverse task requirements.

Despite their effectiveness, these approaches still exhibit limited interactive fusion between visual and textual cues, often failing to direct the model’s attention toward the most relevant visual regions for a given instruction. Qwen-VL~\cite{bai2023qwen} addresses this issue by incorporating cross-attention between image and text tokens, enabling richer cross-modal interaction. However, cross-attention incurs quadratic computational complexity, making it unsuitable for lightweight or latency-constrained deployment.

To overcome these limitations, in Firebolt-VL, we introduce the Cross-Modal Modulator (CMM), which is designed via the integration of a state-space model (SSM)~\cite{gu2024mamba, s4, s4d} and FiLM~\cite{perez2018film} to fuse grid-level visual tokens with textual representations, and efficiently capture the long sequence features for the Large Language Model. By computing lightweight token–grid correlations and applying FiLM-based modulation within an SSM framework, CMM allows the model to dynamically emphasize the most informative visual elements while maintaining near-linear complexity. This design enables stronger fine-grained grounding and contextually accurate multimodal reasoning without the computational overhead of cross-attention, and it can be scaled further for dealing with the long video input or long sequence signal inference of the MLLMs.

%% file: sec/3_method.tex
\begin{figure*}[!t]
    \setlength{\abovecaptionskip}{-2pt}
    \setlength{\belowcaptionskip}{-8pt}
    \centering
    \includegraphics[width=0.98\linewidth]{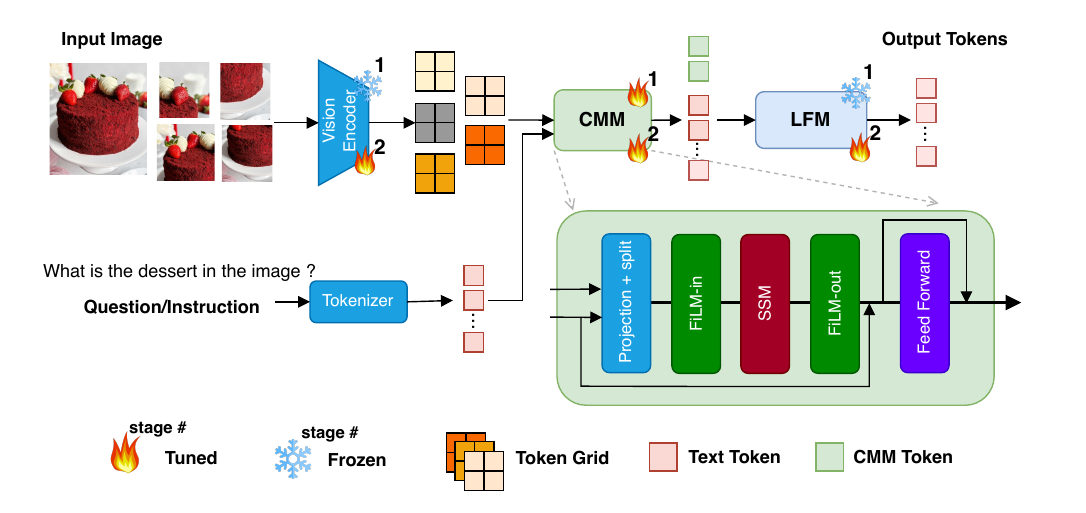}
    \caption{Overview of the Firebolt-VL architecture. The Cross-Modal Modulator (CMM) fuses textual instructions with the visual representations of the query image to produce conditioned tokens, which are then processed by the Liquid Foundation Model (LFM). The model is trained in two stages: (1) CMM pre-training to initialize modulation parameters, and (2) end-to-end training of the full framework.}
    \label{fig:overall}
\end{figure*}

\section{Method}
\subsection{Preliminaries}
\noindent\textbf{State-Space Models (SSM).} An SSM~\cite{hamilton1994state} is a sequence model that modifies the hidden state $\mathbf{h}(t)$ over time through a linear dynamical system.
Following this concept, the Structured State-Space Model (S4)~\cite{s4} and Mamba~\cite{gu2024mamba} introduce a parameterization of $\mathbf{A}$ that guarantees stability and expressiveness, allowing efficient training on long sequences while preserving global dependencies. 
S4 thus bridges the gap between the dynamical-system view of sequence modeling and the content-based attention mechanism \cite{vaswani2017attention} of Transformers.



\noindent\textbf{Feature-wise Linear Modulation (FiLM).} Feature-wise Linear Modulation (FiLM)~\cite{perez2018film} is a lightweight yet effective conditioning mechanism that modulates one modality’s representation based on another by applying learned, feature-wise affine transformations. 
Given a visual feature vector $\mathbf{x} \in \mathbb{R}^{D}$ and a conditioning signal $\mathbf{c}$ (e.g., a text embedding), FiLM generates two modulation parameters, $\gamma(\mathbf{c})$ and $\beta(\mathbf{c})$, through a learnable function such as a linear projection:
\begin{equation}
\mathrm{FiLM}(\mathbf{x}, \mathbf{c}) = \gamma(\mathbf{c}) \odot \mathbf{x} + \beta(\mathbf{c}),
\end{equation}
where $\odot$ denotes element-wise multiplication. 
Through this formulation, FiLM enables the conditioning signal to adaptively scale and shift visual features, integrating semantic cues into the representation without requiring explicit token-level attention.


\subsection{Overview}
Figure~\ref{fig:overall} presents the overall architecture of our framework, which comprises three main modules: the Vision Encoder $Vis(\cdot)$, the Large Language Model $LLM(\cdot)$, and the proposed Cross-Modal Modulator $CMM(\cdot)$.

\subsection{Vision Encoder}

For the vision encoder, we leverage the SigLIP~\cite{siglip2} model, a multilingual vision--language encoder that replaces the traditional softmax contrastive objective with a sigmoid-based loss for image--text alignment. 
Given an input image $X_{I} \in \mathbb{R}^{3 \times H \times W}$, the encoder first divides the image into several patches and processes them through a Vision Transformer (ViT) backbone to obtain global embeddings of grids $X_{v} \in \mathbb{R}^{G \times D_{v}}$, where $G$ denotes the number of visual grids and $D_{v}$ is the embedding dimension. 

In our framework, we employ the SigLIP encoder to extract grid-level visual embeddings $X_{v}$ from the input image while preserving native aspect ratios. 
These embeddings are then projected into the shared latent space for multimodal fusion via the proposed Cross-Modal Modulator (CMM), ensuring fine-grained correspondence between textual cues and spatial visual features.

\subsection{Large Language Model}

A key limitation of Transformer-based MLLMs lies in the quadratic computational complexity of the attention mechanism. Although FlashAttention has been proposed to reduce the complexity in the attention computation, it still poses a challenge due to the KV cache for calculating the attention weight. In this work, we replace the Transformer-based backbone with a Liquid hypothesis-based language model~\cite{hasani2023liquid}, instantiated from the pretrained LFM2 text-only model~\cite{liquid_LFM}. This exploration improves MLLM efficiency as the Liquid-based model achieves substantially smaller time complexity than the attention-based model.

Given the concatenated multimodal representation from CMM and the textual embeddings, denoted as 
$H = [X_{t}; X_{mm}] \in \mathbb{R}^{(L_t +L_{mm}) \times D_{t}}$, 
The model autoregressively generates the target sequence 
$Y = \{y_i\}_{i=1}^{L_y}$ as depicted in Equation~\ref{eq:liquid-autoreg}.
\begin{equation}
    p_{\theta}(Y \,|\, H) = \prod_{i=1}^{L_y} p_{\theta_{i}}(y_i \,|\, H, y_{<i}). \label{eq:liquid-autoreg}
\end{equation}
where $\theta$ denotes all learnable parameters.

Lastly, the predicted tokens are detokenized to produce the final natural-language response. 
By unifying the two modalities with a single autoregressive decoder, Liquid simplifies the architecture, reduces modality-specific alignment overhead, and enables scalable, efficient multimodal reasoning.

\subsection{Cross-Modal Modulator (CMM)}

In prior efficient MLLM works, such as Mobile-VLM~\cite{mobilevlm, mobilevlmv2} and SmolVLM2~\cite{smolvlm}, various projectors have been introduced to align vision-token embeddings with the text-token embedding space. 
Our intuition suggests that the primary challenge of the lightweight Vision-Language Model is the region-level visual awareness, which limits performance on VQA tasks that require rich visual details. To alleviate this limitation, we propose the Cross-Modal Modulator (CMM), which leverages the SSM and FiLM to efficiently attend to the text feature in the specific patches feature. This enables the model to focus more precisely on the task-relevant visual regions to answer the input instruction.

Given the text embedding $X_{t} \in \mathbb{R}^{T \times D_{t}}$, with time step $T$ and the grid-level visual representations $X_{v} \in \mathbb{R}^{G \times D_{v}}$, with the number of grids $G$, the CMM outputs a multimodal representation sequence $X_{mm} \in \mathbb{R}^{T \times D_{t}}$.

To identify the image patches most relevant to each text token, we first compute a correlation matrix between text and vision embeddings. 
Both modalities are projected into a shared latent space as shown in Equation~\ref{equa:project}:
\begin{gather}
    X_{t}^{\prime} = W_{t} X_{t}, \quad X_{v}^{\prime} = W_{v} X_{v} \label{equa:project}
\end{gather}

\noindent The projected embeddings are then split into multiple heads, $X_{t}^{H} \in \mathbb{R}^{H \times T \times D_{H}}$ and $X_{v}^{H} \in \mathbb{R}^{H \times G \times D_{H}}$. 
We compute the scaled dot-product correlation matrix $S_{mm}$ between text tokens and vision grids as:
\begin{gather}
    S_{mm} = \sigma\left(\frac{X_{t}^{H} {X_{v}^{H}}^{\!\top}}{\sqrt{D_H}}\right), \label{equa:corr-score}
\end{gather}
where $\sigma(\cdot)$ denotes the softmax function applied along the grid dimension to normalize the correlation scores to the range $(0,1)$. 
This operation yields attention weights indicating how strongly each text token attends to each visual grid. 
We further retain only the top-$k$ grid locations per token to focus on the most relevant visual patches. 
The optimal $k$ value is analyzed in Section~\ref{subsec:ablation}.

After selecting the relevant patches, we average the correlation matrix across attention heads and compute a weighted sum over the visual features to obtain the per-token visual context embedding $c \in \mathbb{R}^{T \times D_{t}}$ as follows:
\begin{equation}
\begin{aligned}
    \hat{S}_{g} = \frac{1}{H}\sum_{h=1}^{H} S_{mm}^{(h)}, \quad
    c = \sum_{g=1}^{G} \hat{S}_{g} X_{v}^{\prime}. 
\end{aligned}
\end{equation}


\noindent We then leverage the FiLM operator to fuse the text embedding with its corresponding visual context linearly:
\begin{equation}
\begin{aligned}
        X_{f} = \text{LN}(X_t) \odot (1 + \alpha \, \gamma_{\text{in}}) + \alpha \, \beta_{\text{in}}, \label{equa:film} \\ \text{where } [\gamma_{\text{in}}, \beta_{\text{in}}] = W_{f} \, c.
\end{aligned}    
\end{equation}

\noindent Here, $W_{f}$ is the FiLM projection weight mapping the context $c$ into two modulation vectors—the \textit{scale} $\gamma_{\text{in}}$ and the \textit{shift} $\beta_{\text{in}}$. 
The scalar parameter $\alpha$ is a learnable gate that controls the strength of cross-modal modulation. 
Intuitively, $\gamma_{\text{in}}$ scales the feature channels of the text representation based on visual evidence, while $\beta_{\text{in}}$ adds adaptive offsets to introduce new activations conditioned on the image. 
Together, these parameters reshape the text representation according to what each token ``sees'' before sequential modeling.

\noindent The visually modulated representation $X_f$ is then processed by the SSM to capture the hidden-state of the visual-text integration features:
\begin{equation}
    Y_{f} = \text{SSM}(X_{f}). \label{equa:ssm}
\end{equation}

The SSM efficiently models sequential interactions in $\mathcal{O}(T)$ time, allowing the text representation to evolve while preserving visual conditioning. 
Unlike self-attention, which explicitly computes pairwise interactions, the SSM propagates information implicitly through state transitions, capturing both local and global dependencies in a linear and memory-efficient manner. We evaluate alternative SSM variants
in Section~\ref{subsec:ablation}.
\noindent After sequential modeling, we apply a second FiLM modulation, \textit{FiLM-out}, to scale it to the same space of the visual context:

\begin{equation}
    \begin{aligned}
    Y_{f} = Y \odot (1 + \alpha \, \gamma_{\text{out}}) + \alpha \, \beta_{\text{out}}, \label{equa:filmout} \\ \text{where }
    [\gamma_{\text{out}}, \beta_{\text{out}}] = W_{f}^{\prime} \, c.
\end{aligned}
\end{equation}

\noindent Similar to FiLM-in, $\gamma_{\text{out}}$ and $\beta_{\text{out}}$ adjust the post-SSM features based on $c$, ensuring alignment between textual and visual features after temporal mixing.

\noindent Finally, residual connections and a feed-forward network (FFN) refine the fused representations:
\begin{equation}
    X_{\text{out}} = \text{LN}\big((X_t + Y_{f}) + \text{FFN}(X_t + Y_{f})\big), \label{equa:ffn}
\end{equation}
\noindent where LN denotes the Layer Normalization, and FFN represents the Feed Forward layer. This step stabilizes the multimodal representation and enhances expressiveness through non-linear transformations in the FFN. 
The output $X_{\text{out}} \in \mathbb{R}^{B \times T \times D_t}$ thus contains text features that are visually modulated and temporally contextualized by the SSM.

\noindent To obtain a global multimodal representation, we aggregate the token-level outputs $X_{\text{out}}$ using mean pooling:
\begin{equation}
    z = \frac{1}{T} \sum_{t=1}^{T} X_{\text{out},t}. \label{equa:pooling}
\end{equation}

The resulting vector $X_{mm} \in \mathbb{R}^{B \times D_t}$ serves as a compact fused embedding that captures both linguistic and visual semantics. 
This embedding is subsequently passed to the language decoding stage for multimodal reasoning and response generation.

\noindent \textbf{Complexity Analysis:} CMM performs the token-grid correlation only once to derive a compact visual context and replaces the repeated cross-attention mechanism with a linear-time State-Space Model (SSM), whose complexity scales as $\mathcal{O}(T G D_t + T D_t^2 + T D_t f(T))$, where $f(T) \in \{1, \log T\}$. As a result, CMM achieves nearly linear scaling with respect to sequence length $T$ and cost with respect to grid size $G$ (for fixed or sparse top-$k$), substantially reducing both computational and memory overhead while preserving effective cross-modal alignment through FiLM-based modulation. In contrast, typical multimodal fusion relies on cross-attention between text and vision tokens, which computes the attention matrix $S_{mm} = \text{Softmax}(QK^{\top}/\sqrt{D_t})$ and the weighted aggregation $AV$, resulting in a computational complexity of $\mathcal{O}(T G D_t)$ per layer and a memory requirement proportional to $T \times G$. While effective, this operation becomes expensive as the scaling of the text length $T$ and visual grids $G$ when compared with CMM.

%% file: sec/4_experiments.tex
\begin{table*}[ht]
\small
    \centering
 \resizebox{0.95\textwidth}{!}{
    \begin{tabular}{l|c|c|ccccccc}
    \hline
       \textbf{Method}  & \textbf{LLM} & \textbf{Parameters} & \textbf{VQAv2} & \textbf{POPE} & \textbf{AI2D} & \textbf{MMMU$_{\text{val}}$} & \textbf{MME$^{p}$} & \textbf{SQA$^{I}$} & \textbf{MMB$_{\text{dev}}$} \\
    \hline
    IDEFICS~\cite{laurenccon2023obelics} & LLaMA & 9.0B & 60.0 & 81.9 & 42.2 & 18.4 & 1177.3 & 	
53.5 & 45.3 \\
      OpenFlamingo v2~\cite{openflamingo} & MPT & 9.0B & 60.4 & 52.6 & 31.7 & \underline{28.8} & 607.2 & 44.8 & -- \\
        MiniGPT-4-v1~\cite{zhu2023minigpt} & Vicuna & 8.0B & -- & 34.6 & 28.4 & 23.6 & 1047.4 & 	
39.6 & -- \\
    MiniGPT-4-v2~\cite{chen2023minigpt} & LLaMA 2 & 8.0B & -- & 60.0 & 30.5 & 25.0 & 968.4& 54.7 & -- \\  
    Chameleon~\cite{team2024chameleon} & LLaMA 2 & 7.0B & -- & 19.4 & \underline{46.0} & 22.4 & 202.7 & 46.8 & -- \\
    FastV~\cite{chen2024image}  & Vicuna & 7.0B & 55.0 & 48.0 & 42.7 & 22.0 & 873.2 & 51.1 & -- \\

    \hline 
        Kosmos-2~\cite{peng2023kosmos} & GPT 2 & 1.7B & 45.6 & 66.3 & 25.6 & 23.7 & 721.1 & 32.7& -- \\
       MobileVLM~\cite{mobilevlm} & Mobile-LLaMA & 1.7B  & -- & 84.5 & 36.6 & 25.3 & 1196.2 & 57.3 & 59.6\\
       MobileVLM V2~\cite{mobilevlmv2} & Mobile-LLaMA & 1.7B & -- & 84.3 & 38.1 & 19.0 & 1302.8 & \underline{66.7} & 57.7\\
       MoE-LLaVA~\cite{llava-moe} & Qwen  & 2.2B & \underline{76.2} & \textbf{87.0}  & 42.1 & 26.6 & 1291.6 & 63.1 & 59.6 \\
        LLaVA-Phi~\cite{llavaphi} & Phi 2 & 2.7B & 71.4 & \underline{85.0}  & -- & -- & \underline{1335.1} & \textbf{68.4} & \underline{59.8} \\
        
        SmolVLM2~\cite{smolvlm}  & Mobile-LLaMA & 0.3B & -- & 	
54.3 & 39.2 & \textbf{28.9} & 1236.5 & 58.8 & --\\
        \hline
        \rowcolor{lightash}
        \textbf{Firebolt- VL (Ours)} & \textbf{LFM2} & 0.8B & \textbf{76.6} & 69.4 & \textbf{46.2} & 26.4 & \textbf{1376.2} & 56.7 & \textbf{64.6} \\
        \hline
    \end{tabular}
    }
    \caption{Quantitative comparison of the proposed \textbf{Firebolt-VL} model with existing MLLMs across seven benchmarks. The superscript $p$ denotes the perception score on the MME benchmark, while SQA$^{I}$ refers the ScienceQA-IMG~\cite{sciqa}. The best results are shown in \textbf{bold}, and the second-best results are \underline{underlined}. “--” indicates results not reported in the original papers.}
    \label{tab:benchmark_firebolt}
\end{table*}

\section{Experimental Setup}
\subsection{Training Recipe}

We train the Firebolt-VL model in two stages, as illustrated in Figure~\ref{fig:overall}. In the first stage, we initialize the CMM module while freezing the vision encoder and the language model. The CC3M dataset~\cite{cc3m1, cc3m2} is employed for the initialization. In the second stage, we perform end-to-end training to enhance its reasoning ability. Specifically, we leverage the LLaVA-CoT dataset~\cite{llavacot}, and our processed MMPR-v1.2~\cite{mmpr1, mmpr2, mmpr3} dataset follows the chain-of-thought format of the LLaVA-CoT to enable the model to learn reasoning capabilities.

\subsection{Implementation Details}
Both stages of the model are trained using 2 NVIDIA H100 80GB GPUs with a batch size of 128 in stage 1, and 8 in stage 2. For the optimizer, we employ the AdamW optimizer with a learning rate of $5\times10^{-4}$ for the first stage and $1\times10^{-4}$ for the second stage. The model is trained for 5 epochs in each of the two stages. The best model is selected after 2 epochs in each stage. We select the best model using the perplexity metric on the validation set. 

\subsection{Comparison Baseline}
To assess the generalization and reasoning ability of Firebolt-VL across diverse environments, we evaluate on multiple benchmarks such as VQAv2~\cite{vqav2}, POPE~\cite{pope}, AI2D~\cite{ai2d}, MMMU~\cite{yue2024mmmu} validation set, MME~\cite{mme}, SQA-Image~\cite{sciqa}, and MMB \cite{liu2024mmbench} development set.  For a fair comparison, we conduct the benchmark against models trained on a comparable dataset scales. We consider two evaluation settings. The first setting compares against the big models, which have more than 7 billion parameters, including IDEFICS (2023)~\cite{laurenccon2023obelics}, OpenFlamingo v2 (2023)~\cite{openflamingo}, MiniGPT-4-v1 (2023)~\cite{zhu2023minigpt}, MiniGPT-4-v2 \cite{chen2023minigpt},  Chameleon (2024) \cite{team2024chameleon}, and FastV (2024)~\cite{chen2024image}. The second setting compares small models, which have fewer than 3 billion parameters, involving these methods: Kosmos-2 (2023)~\cite{peng2023kosmos}, MobileVLM (2024)~\cite{mobilevlm}, MobileVLM V2 (2024)~\cite{mobilevlmv2}, MoE-LLaVA (2024)~\cite{llava-moe}, LLaVA-Phi (2024)~\cite{llavaphi}, and SmolVLM2 (2025)~\cite{smolvlm}.

\section{Results}
\begin{figure*}[!ht]
  \setlength{\abovecaptionskip}{-2pt}
  \setlength{\belowcaptionskip}{-4pt}
  \centering
  \makebox[\linewidth][c]{%
    \hspace{-0.02\linewidth}
    \includegraphics[width=0.92\linewidth]{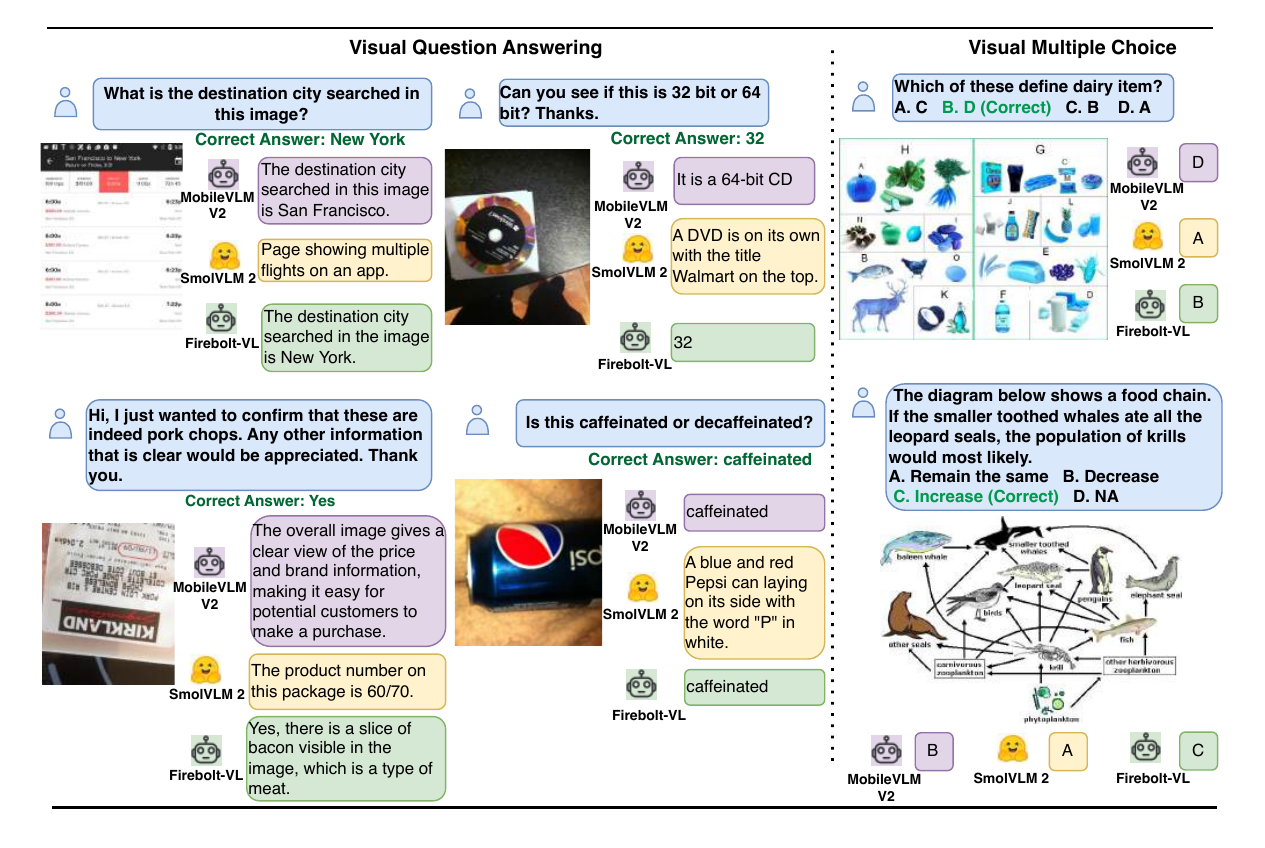}
  }
  \caption{Qualitative comparison of responses from Firebolt-VL with recent efficient vision--language models, including MobileVLM V2~\cite{mobilevlmv2} and SmolVLM2~\cite{smolvlm}, on detail-dependent question-answering tasks. Firebolt-VL demonstrates stronger fine-grained grounding and more accurate, instruction-aligned responses.}
  \label{fig:firebolt_vl_responses}
\end{figure*}
\subsection{Quantitative Results}

\noindent\textbf{Comparison with previous works in image understanding.}
From the benchmark results in Table~\ref{tab:benchmark_firebolt}, despite having fewer than 1B parameters, Firebolt-VL achieves competitive performance compared to much larger models (over 7B parameters), and clearly outperforms compact models in the 0.3B–3B range. Although there is a performance drop on the POPE benchmark, this may be due to the SSM's implicit state propagation being less effective at suppressing hallucinated content than explicit attention. Notably, Firebolt-VL attains the highest scores on both VQAv2 and MME benchmarks, demonstrating strong visual reasoning and perceptual alignment capabilities, which reflect the efficacy of the CMM module in enhancing the perception ability of the MLLMs. These findings highlight the effectiveness of our designed fusing module in capturing long multimodal sequence dependencies efficiently, without relying on a heavy attention-based mechanism. Results suggest that structured state-space modeling offers a promising alternative for scalable and efficient multimodal understanding.

\begin{figure}[!ht]
    \centering
    \setlength{\abovecaptionskip}{-1pt}
    \includegraphics[width=0.75\linewidth]{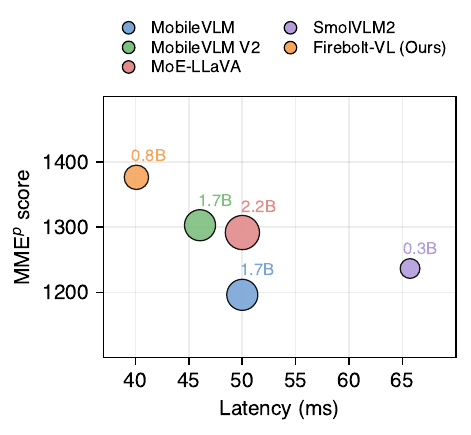}
    \caption{Accuracy--latency comparison of compact MLLMs. Bubble area denotes parameter count (B), and annotations indicate the exact model size. Firebolt-VL provides a favorable accuracy--latency trade-off, achieving a higher MME$^{p}$ perception score at lower inference latency.}
    \label{fig:perception_vs_accuracy_MMLM}
\end{figure}

\noindent \textbf{Comparison of efficiency with prior efficient vision-language models.}
To evaluate the computational efficiency of Firebolt-VL against MobileVLM~\cite{mobilevlm}, MobileVLM V2~\cite{mobilevlmv2}, MoE-LLaVA~\cite{llava-moe}, and SmolVLM2~\cite{smolvlm}, we conduct experiments on the POPE~\cite{pope} dataset using two key metrics: \textit{latency} and \textit{throughput} (tokens per second). All models were evaluated on a single NVIDIA H100 GPU with a maximum output length of 256 tokens for consistent comparison. As shown in Figure~\ref{fig:eff_throughput}, Firebolt-VL achieves the highest throughput at 46.67 (tokens/sec) among lightweight multimodal baselines. Figure~\ref{fig:perception_vs_accuracy_MMLM} further illustrates the accuracy–latency trade-off, where Firebolt-VL achieves competitive perception performance while maintaining low latency. These results demonstrate the strong computational efficiency of our Liquid-based backbone and the lightweight nature of our Cross-Modal Modulator. Overall, the integration of structured state-space modeling enables fast multimodal reasoning while maintaining low computational overhead, making Firebolt-VL suitable for real-time and resource-constrained deployment scenarios.



\subsection{Qualitative Results}
In Figure~\ref{fig:firebolt_vl_responses}, we present qualitative results that highlight Firebolt-VL’s visual reasoning and text generation abilities across both visual question answering and visual multiple-choice tasks, compared against two efficient multimodal baselines—SmolVLM2~\cite{smolvlm} and MobileVLM V2~\cite{mobilevlmv2}. Unlike prior lightweight models, which often produce generic descriptions, Firebolt-VL consistently generates precise, question-grounded answers. For example, when asked about the “destination city being searched,” Firebolt-VL correctly attends to the search bar region in the image and extracts the appropriate answer, while baselines fail to localize this detail.

\noindent Similarly, in multiple-choice question, Firebolt-VL demonstrates reliable fine-grained visual discrimination, such as identifying subtle differences among dairy product labels or tracking hierarchical relations in food chains. These examples collectively show that Firebolt-VL effectively attends to task-relevant visual grids and leverages localized visual cues to produce more accurate and context-aware responses than existing efficient VLMs. Therefore, these results showcase the perceptual improvements introduced by the CMM module, which strengthens the overall framework and can then be extended for further applications.

\subsection{Ablation Studies}
\label{subsec:ablation}
\noindent \textbf{State-space model (SSM) choice.} To determine the most suitable state-space model for our Firebolt-VL framework, we conduct experiments to evaluate the impact of different SSM variants on overall performance. Specifically, we compared three representative models—Mamba~\cite{gu2024mamba}, S4D~\cite{s4d}, and S4~\cite{s4}, as depicted in Table~\ref{tab:ssm_choice}.

\begin{table}[!ht]
\centering
\begin{tabular}{lccc c}
\toprule
\textbf{Approach} & \textbf{POPE} & \textbf{AI2D} & \textbf{MMMU}$_\text{val}$ & \textbf{Average} \\
\midrule
Mamba~\cite{gu2024mamba} & 57.4 & 39.9 & 23.6 & 40.3 \\
S4D~\cite{s4d} & \textbf{69.9} & 45.6 & 23.7 & 46.4 \\
\textbf{S4}~\cite{s4} & 69.4 & \textbf{46.2} & \textbf{26.4} & \textbf{47.3} \\
\bottomrule
\end{tabular}
\caption{Performance comparison of Mamba, S4D, and S4 state-space models on POPE~\cite{pope}, AI2D~\cite{ai2d}, and MMMU$_\text{val}$~\cite{yue2024mmmu} benchmarks. Structured models (S4/S4D) outperform Mamba.}
\label{tab:ssm_choice}
\end{table}
From the results, it can be observed that S4 and S4D yield higher performance compared to Mamba, indicating that structured state-space models are more effective in capturing the interactions between textual and visual embeddings. Since the visual embeddings represent information from five spatial grids of the image, the structured state-space architecture enables more accurate modeling of geometric relationships during the transition process, thereby achieving superior results compared to the Mamba model.

\noindent \textbf{Cross-modal fusion mechanism choice.} To evaluate the effectiveness of the CMM module, we conduct experiments comparing three fusion strategies: (1) Prepend, where the projected image features from the MLP layer are passed through the language Model; (2) Cross-attend, where cross-attention follows Q-Former implementation~\cite{li2023blip} is applied to enable interaction between visual and textual features; and (3) CMM (ours), the proposed module that integrates state-space modeling for efficient and structured cross-modal fusion. The experimental results are depicted in Table~\ref{tab:connector}.

\begin{table}[!ht]
\centering
\begin{tabular}{lccc}
\toprule
\textbf{Approach} & \textbf{MME} & \textbf{AI2D} & \textbf{MMMU}$_\text{val}$ \\
\midrule
Prepend & 981.5 & 24.1 & 22.1 \\
Cross-attended & 1036.5 & 45.1 & 24.4 \\
\rowcolor{lightash}
\textbf{CMM (Ours)} & \textbf{1376.2} & \textbf{46.2} & \textbf{26.4} \\
\bottomrule
\end{tabular}
\caption{Performance of different connector methods on MME~\cite{mme}, AI2D~\cite{ai2d}, and MMMU$_\text{val}$~\cite{yue2024mmmu}. Our proposed CMM outperforms both Prepend and Cross-attention fusion across all benchmarks.}
\label{tab:connector}
\end{table}

From the benchmark results, we observe that in the MME perception benchmark, the CMM approach significantly outperforms both the Cross-attend and Prepend methods, demonstrating that our fusion mechanism effectively enhances the model’s perceptual ability. Moreover, in the AI2D and MMMU$_{val}$ benchmarks, CMM also achieves higher scores—particularly on chart and diagram questions in AI2D and on mathematics and coding-related questions in MMMU$_{val}$. In detail, it shows that the combination of FiLM and state-space can effectively fuse the vision and the text features, where the text features are long-range dependencies when compared with the Q-Former approach, showing a considerable improvement in the benchmark results. In general, these results indicate that CMM enables more precise contextual alignment between modalities, leading to an overall improvement in the model’s reasoning and understanding performance.

\noindent \textbf{Top-$k$ grid assessment.}
To assess how many visual grids should be selected by the CMM module, we evaluate model performance under different values of $k$, where $k$ represents the number of top-ranked visual grids processed by the fusion mechanism. Due to computational constraints, the vision encoder produces five grids; therefore, we evaluate top-$k$ values ranging from 1 to 5.

\begin{table}[!ht]
\setlength{\belowcaptionskip}{-2pt}
\centering
\small
\setlength{\tabcolsep}{5pt}
\renewcommand{\arraystretch}{1.12}
\begin{tabular}{lccccc}
\toprule
\textbf{Dataset} & \textbf{$k=1$} & \textbf{$k=2$} & \textbf{$k=3$} & \textbf{$k=4$} & \textbf{$k=5$} \\
\midrule
\textbf{MME} & 1039.8 & 1192.0 & 1258.6 & \textbf{1376.2} & 1289.4 \\
\textbf{AI2D} & 43.9 & 44.1 & 45.4 & \textbf{46.2} & 45.6 \\
\textbf{MMMU}$_\text{val}$ & 22.2 & 24.3 & 25.8 & \textbf{26.4} & 26.0 \\
\bottomrule
\end{tabular}
\caption{Effect of top-$k$ grid selections on MME~\cite{mme}, AI2D~\cite{ai2d}, and MMMU$_\text{val}$~\cite{yue2024mmmu}. Performance improves up to $k=4$ and slightly drops at $k=5$.}
\label{tab:topk}
\end{table}

\noindent From Table~\ref{tab:topk}, we observe that performance improves consistently as $k$ increases from 1 to 4 across all benchmarks (MME, AI2D, and ${\text{MMMU}_\text{val}}$). Notably, the results on MMMU$_\text{val}$ increase as top-$k$ increases, suggesting that reasoning-intensive tasks benefit from integrating multiple visual cues. These results indicate that relying solely on the single most salient grid ($k=1$) is insufficient for robust multimodal understanding, and that attending to multiple top-ranked grids allows the model to capture fine-grained details and complementary visual signals better. However, when $k=5$, there is a slight performance drop. Overall, increasing $k$ allows the model to aggregate information from a broader visual context, which may help capture additional details that are relevant to the question; however, selecting too many grids can also introduce noise or dilute the contribution of the most informative regions.


%% file: sec/5_conclusion.tex
\section{Conclusion}

We present Firebolt-VL, an efficient multimodal LLM that leverages the Liquid-based language decoder and incorporates a lightweight fusion mechanism combining a state-space model with linear feature modulation. This design specifically addresses the challenge of fine-grained detail perception in efficient vision–language models. We demonstrate that our proposed Cross-Modal Modulator (CMM) enables the model to be aware of the precise visual details that are directly relevant to the input question or instruction. By integrating a Liquid-based language model (LFM2)~\cite{liquid_LFM} within the VLM framework, Firebolt-VL achieves competitive performance compared to attention-based architectures while maintaining a lightweight design. As a result, our approach achieves strong perceptual and reasoning performance with reduced computational overhead. This approach can potentially be extended to long video sequences and broader visual grounding tasks.

\noindent\textbf{Limitations.} Our current model operates on single-image inputs, and extending CMM to support multiple images or video sequences remains an open challenge. We have not yet developed a version of CMM optimized for video inputs; however, future work will explore efficient cross-modal connectors capable of integrating temporal visual information with textual instructions.

\section*{Acknowledgements}
This work was supported by the Research Council of Finland (Grant No. 362729), Business Finland (Grant No. 169/31/2024), and the Finnish Doctoral Program Network in Artificial Intelligence, AI-DOC (Grant No. VN/3137/2024-OKM-6), awarded to Bo Zhao.
  D. Jha is supported in part by the U.S. Department of Education (P116Z240151 to the University of South Dakota and the South Dakota School of Mines \& Technology). The views expressed are those of the author(s) and do not necessarily represent the official views of the U.S. Department of Education.